\documentclass[10pt,twocolumn,letterpaper]{article}

\usepackage{cvpr}

\usepackage{cite}
\usepackage{amsmath,amssymb,amsfonts}
\usepackage{algorithmic}
\usepackage{graphicx}
\usepackage{textcomp}
\usepackage{epsfig}
\usepackage{multirow}
\usepackage{booktabs}
\usepackage{times}
\usepackage{graphics}
\usepackage{enumitem}
\usepackage{caption}
\usepackage{threeparttable}
\usepackage[breaklinks=true,bookmarks=false]{hyperref}
\cvprfinalcopy % *** Uncomment this line for the final submission

\def\cvprPaperID{****} % *** Enter the CVPR Paper ID here

\begin{document}

%%%%%%%%% TITLE
\title{LoViT: Long Video Transformer for Surgical Phase Recognition
}
\author{Yang Liu$^1$,\and Maxence Boels$^1$, \and Luis C. Garcia-Peraza-Herrera$^1$,\and Tom Vercauteren$^1$,\and Prokar Dasgupta$^2$, \and Alejandro Granados$^1$
,\and S\'{e}bastien Ourselin$^1$
\and\\[2mm]
$^1$Department of Surgical \& Interventional Engineering, King's College London\\
    $^2$Department of Peter Gorer Department of Immunobiology, King's College London\\[2mm]
E-mail: yang.9.liu@kcl.ac.uk\\
}

\maketitle
%\thispagestyle{empty}
%\footnote{The work of Yang Liu~(e-mail:yang.9.liu@kcl.ac.uk) is supported by China Scholarship Council under Grant 202106160011.
%Yang Liu, Maxence Boels, Luis C. Garcia-Peraza-Herrera, Tom Vercauteren, Alejandro Granados, and S\'{e}bastien Ourselinis are with the  Prokar Dasgupta is with the }

%%%%%%%%% ABSTRACT

\begin{abstract}
% Context
Online surgical phase recognition plays a significant role towards building contextual tools that could quantify performance and oversee the execution of surgical workflows.
% Motivation
Current approaches are limited since they train spatial feature extractors using frame-level supervision that could lead to incorrect predictions due to similar frames appearing at different phases, and poorly fuse local and global features due to computational constraints which can affect the analysis of long videos commonly encountered in surgical interventions.
In this paper, we present a two-stage method, called Long Video Transformer (LoViT) for fusing short- and long-term temporal information that combines a temporally-rich spatial feature extractor and a multi-scale temporal aggregator consisting of two cascaded L-Trans modules based on self-attention, followed by a G-Informer module based on \textit{ProbSparse} self-attention for processing global temporal information. The multi-scale temporal head then combines local and global features and classifies surgical phases using phase transition-aware supervision.
Our approach outperforms state-of-the-art methods on the Cholec80 and AutoLaparo datasets consistently. Compared to Trans-SVNet, LoViT achieves a 2.4 pp~(percentage point) improvement in video-level accuracy on Cholec80 and a 3.1 pp improvement on AutoLaparo.
%Moreover, it achieves a 5.2 pp improvement in phase-level Jaccard on AutoLaparo and a 1.7 pp improvement on Cholec80.
Our results demonstrate the effectiveness of our approach in achieving state-of-the-art performance of surgical phase recognition on two datasets of different surgical procedures and temporal sequencing characteristics whilst introducing mechanisms that cope with long videos. The code will be available at \url{https://github.com/MRUIL/LoViT}
\end{abstract}

\section{Intruduction}

Surgical Data Science (SDS) aims to improve the quality of interventional healthcare through the capture, modelling, and analysis of patient data from medical devices within the operating room (OR)~\cite{DBLP:journals/corr/abs-2011-02284}. Surgical phase and action recognition are paramount in comprehending surgical processes, evaluating surgeon performance, and providing assistance that is reactive to the surgical context~\cite{DBLP:journals/pieee/VercauterenUPN20}. Specifically, automatic recognition of surgical phases and actions plays a crucial role in developing surgical skills and enhancing the efficiency and safety of surgeries by providing immediate feedback to the surgical team. 
%Additionally, it enables the development of surgical education systems that facilitate the training and testing of surgical trainees. 
This is critical in advancing surgical practices and education through continuous improvement.
During endoscopic-based interventions, surgical phase recognition aims to classify every video frame into high-level stages of surgery~\cite{garrow2021machine}, while action recognition aims to classify each frame into granular and fine tasks entirely from data. 
% Not sure if you need the sentence below, unless you introduce first online and offline annotation and describe also offline.
In contrast to action recognition, surgical phase recognition requires approaches that process videos over long time frames since each phase typically contains several actions. Compared to offline recognition that can be used for automated annotation of prerecorded videos, online recognition allows recognizing current activity without future information, which could be used to alert surgeons of those tasks that are likely to lead to complications~\cite{DBLP:journals/tmi/QuellecLCC15} and support their decision making~\cite{DBLP:journals/cars/DergachyovaBHMJ16}.
Early work related to surgical phase recognition focused mostly on statistical models.
%{Earliest attempts for surgical phase recognition mainly adopted statistical models.}
Blum~\etal~\cite{DBLP:conf/miccai/BlumFN10} analyzed surgical signals, including the use of surgical instruments and high-frequency coagulation and cutting, for dimensionality reduction of video frames into image features.
Padoy~\etal~\cite{DBLP:journals/mia/PadoyBAFBN12} processed synchronized signals based on Dynamic Time Warping~\cite{Dynamic_1978} and Hidden Markov Models(HMM)~\cite{DBLP:journals/pieee/Rabiner89}.  Bardram~\etal~\cite{DBLP:conf/percom/BardramDJLNP11} presented a sensor platform and a machine learning approach to sense surgical phases. Holden~\etal~\cite{DBLP:journals/tbe/HoldenUSMCGPF14} developed a workflow segmentation algorithm for needle interventions using needle tracking data. 
% Q: what do you mean manual annotation? is that for ground truth generation? something else? and for equipment installation, do you mean for sensors?
% Q: what do you mean by 'pure' video? 
% manual annotation is still a limitation of current approaches ...
However, these techniques exploit extra information, such as manual instrument annotation or equipment installation, that might be inconvenient to collect and could cause workload, rather than only video. 
%However, it's inconvenient to collect those extra auxiliary information and cause a greater workload in a real scene, which is contrary to our original intention. 
Pure video-based methods were then proposed to overcome the aforementioned limitations.
Quellec~\etal~\cite{DBLP:journals/tmi/QuellecLCC15} presented a multiscale motion characterization approach based on adaptive spatio-temporal polynomials.% \added[id=AG]{for ...}. 
~Dergachyova~\etal~\cite{DBLP:journals/cars/DergachyovaBHMJ16} trained a set of \textit{AdaBoost} classifiers capable of distinguishing one surgical phase from others, and then adopted a hidden semi-Markov model to obtain a final decision. 
% Q: please check. it is not very clear what pre-defined dependencies mean? A: just copy it from somewhere else, I think it means those methods define some relationships and features manually
Since vision features and temporal context of surgical video are highly complex, these methods show limited representation capabilities with pre-defined dependencies~\cite{DBLP:journals/tmi/JinD0YQFH18,DBLP:conf/miccai/GaoJLDH21}.

% I would probably say recently for a year or two.
% what do you mean by "pure video"? only video frames? Yang: Yes
With the advent of deep learning, approaches entirely based on video frames as input data were proposed.
%{Recently, some learning-based methods have been proposed.}
EndoNet~\cite{DBLP:journals/tmi/TwinandaSMMMP17} was the first work to use a Convolutional Neural Network (CNN)~\cite{CNN_1998} for multiple recognition tasks, which extracted spatial features that were fed into a hierarchical HMM for modeling temporal information.
Twinanda~\cite{DBLP:phd/hal/Twinanda17} used Long Short-term Memory (LSTM)~\cite{LSTM_1997}, a type of Recurrent Neural Network (RNN) gate, rather than an HMM to enhance EndoNet's ability to understand the temporal context.
Jin~\etal~\cite{DBLP:journals/mia/JinLDCQFH20} presented a multi-task framework called MTRCNet-CL, which used a correlation loss to exploit the relatedness between tool presence detection and surgical phase recognition with the aim of simultaneously boosting the performance of both tasks. 
% I think we need to be clear early in this paper that we are covering supervised methods only since you also work with phase labels
Similar to early work, these methods are based on multi-task learning which necessitates additional tool annotations in addition to phase labels, thus increasing workload and costs. To mitigate this, the following methods have shifted focus to solely learning a single task using only phase labels.
SV-RCNet~\cite{DBLP:journals/tmi/JinD0YQFH18} leveraged a Residual network (ResNet)~\cite{resnet_he} and an LSTM network to learn spatial and temporal features end-to-end, respectively. 
% Q: what are hard frames? please define.
Yi~\etal~\cite{DBLP:conf/miccai/YiJ19} presented an Online Hard Frame Mapper (OHFM) to handle the detected hard frames, that defined as some frames with indistinguishable visual features but are separately assigned with different labels. Gao~\etal~\cite{DBLP:conf/icra/GaoJDH20} designed a framework using a tree search algorithm to consider future information from LSTM. 
However, the ability of LSTMs to retain memory is limited due to their vanishing gradients problem, which affects their ability to capture information from long sequences. This particularly affects their use on surgical videos for temporal feature extraction since interventions could last a few hours~\cite{DBLP:conf/miccai/CzempielPKSFKN20}. In contrast, TMRNet~\cite{DBLP:journals/tmi/JinLCZDH21} employed a non-local bank operator to establish the relationship between the current frame and all previous features generated by LSTM. However, the contribution of all previous features to the current frame is only a simple weighted sum, resulting in a lack of effective capture of global information.
Czempiel~\etal~\cite{DBLP:conf/miccai/CzempielPKSFKN20} presented TeCNO, based on Temporal Convolutional Networks (TCNs)~\cite{DBLP:conf/eccv/LeaVRH16, DBLP:conf/cvpr/FarhaG19}, that is able to capture long-term temporal correlations. 
However, by essentially adapting dilated convolutions~\cite{DBLP:conf/ssw/OordDZSVGKSK16} for long sequences using TCNs, the increased receptive field obtained through dilation can result in a loss of fine-grained relationships between more distant time steps. In other words, the information captured by TCNs becomes coarser as the sequence length increases. This can result in a loss of detail in the learned representations and reduce performance.
% Moving beloiw since this already talkes about Transformers that you will introduce later on
%\deleted[id=AG]{Even Trans-SVNet~\cite{DBLP:conf/miccai/GaoJLDH21} tried to search lost information through tailing a small Transformer model to fuse temporal feature with previous spatial features, this problem is still exist.}

With the rise of Transformers in Natural Language Processing (NLP)~\cite{DBLP:conf/nips/VaswaniSPUJGKP17}, the parallel nature of attention blocks were quickly used to replace CNNs and RNNs architectures. In vision-related tasks, Vision Transformer (ViT)~\cite{DBLP:conf/iclr/DosovitskiyB0WZ21} was proposed by replacing word tokens of vanilla Transformers with sequences of $16\times16$ pixels image patches.
Temporal attention was then introduced with TimeSformer~\cite{gberta_2021_ICML} and VideoViT~\cite{DBLP:conf/iccv/Arnab0H0LS21} approaches for video classification with self-attention on sequences of frame-level patches. 
% You have been using AVT for longer times, is the claim for working only for over one minute long videos claim correct? A: I only use a clip to train AVT, similar with using frame-level supervision to train ResNet
% TODO could you please be precise on how many dozens are used in this AVT approach? Y:(Done)
The Anticipative Video Transformer (AVT)~\cite{DBLP:conf/iccv/GirdharG21} then ranked first in EPIC-Kitchens-100 Challenge~\cite{Damen2022RESCALING} by using a ViT backbone and a causal head layer to predict the next actions with the disadvantage of working only on videos being dozens of frames long.
Czempiel~\etal~\cite{DBLP:conf/miccai/CzempielPOKBN21} first introduced a Transformer-based temporal future extractor for surgical phase recognition. This was then followed by Trans-SVNet~\cite{DBLP:conf/miccai/GaoJLDH21}, which aimed to address the issue of lost fine-grained information in TCNs by using a small Transformer model to complement the spatial and temporal features that were previously ignored. Despite this attempt, the drawback of the dilated convolution structure in TCNs still remains. This problem exacerbates since time and memory complexity of Transformers are quadratic due to their dot-product operations, a particular problem for long videos commonly resulting from surgical interventions.
Informer~\cite{DBLP:conf/aaai/ZhouZPZLXZ21} was proposed to overcome the limitations of Transformers when processing long sequences. It utilizes attention mechanisms to accurately capture long-range dependencies within the sequences. To address the inherent problems of Transformers, such as their quadratic time complexity and high memory usage of $\mathcal{O}(L^2)$~(where $L$ means the input sequence length), Informer introduces a novel \textit{ProbSparse} self-attention mechanism that reduces the time and memory complexity to $\mathcal{O}(L \log(L))$.

%While today's video recognition systems can process snapshots or short clips accurately, they are unable to learn spatial and temporal representations across longer time spans. State-of-the-art architectures \added[id=AG]{typically process videos at 1~Hz and} can only process less than 5 seconds of a video without hitting the computation and memory bottlenecks of computing infrastructure. Long videos with high frame rate (HFR) are computationally expensive as they require more time to process and memory on hardware. This causes an increase in processing time and memory requirements. Hence, performing surgical phase recognition during long surgeries\added[id=AG]{, whilst keeping fine-grained temporal information,} can quickly become difficult to perform at scale.

Motivated by designing deep learning algorithms that can efficiently interpret surgical phases of long videos without impacting performance, we propose \textbf{LoViT} -- a Long Video Transformer -- that outperforms state-of-the-art work. The specific contributions of this work include: 
\begin{itemize}
    \item a temporally-rich spatial feature extractor that leverages temporal information to improve the precision of spatial feature learning,
    \item multiscale temporal feature aggregation of local information using vanilla self-attention, and global relationships following \textit{ProbSparse} self-attention, and
    \item a simple but effective phase transition-aware supervision for highlighting critical temporal information of phase transitions.
\end{itemize}

\section{Methods}

In this work, we target the problem of online surgical phase recognition. Formally, this is a video classification problem where we aim to solve for a mapping $f$ such that 
$f_{\boldsymbol{\theta}} \left( X_t \right) \approx p_t$, 
where $X_t = \{ \boldsymbol{x}_j \}_{j = 1}^{t}$ is a given input video stream, and
$\boldsymbol{x}_j \in \mathbb{R}^{H\times W \times C}$. 
The symbols $H$, $W$, $C$ represent the image height, width, and number of channels, respectively. 
As in our work we deal with RGB images, $C=3$. 
The height and width of each video frame change from dataset to dataset.
The first frame of the video is noted as $\boldsymbol{x}_1$, and the current $t$-th frame as 
$\boldsymbol{x}_t$.
The output $p_t \in \{ k \}_{k=1}^{K}$ is the class index corresponding to the surgical phase of the video frame $\boldsymbol{x}_t$, where $K$ is the total number of classes or surgical phases. 
The symbol $\boldsymbol{\theta}$ is a vector of parameters corresponding to the weights of our network model $f$, which we call LoViT throughout the paper.

%For a surgical video, represented by the sequence $X = (x_1, x_2, ..., x_n)$, where $x_i\in \mathbb{R}^{H\times W \times C}$ denotes the $i$-th image frame, LoViT is designed to identify the current surgical phase $p_i$ at each frame $x_i$ from $N_p$ possible classes, based only on the information from previous frames $X_i = x_{1:i}$, which can be expressed mathematically as $\hat{p_i} = \rm{LoViT} (X_i)$.

\begin{figure*}[ht]
\begin{center}
\includegraphics[width=\linewidth]{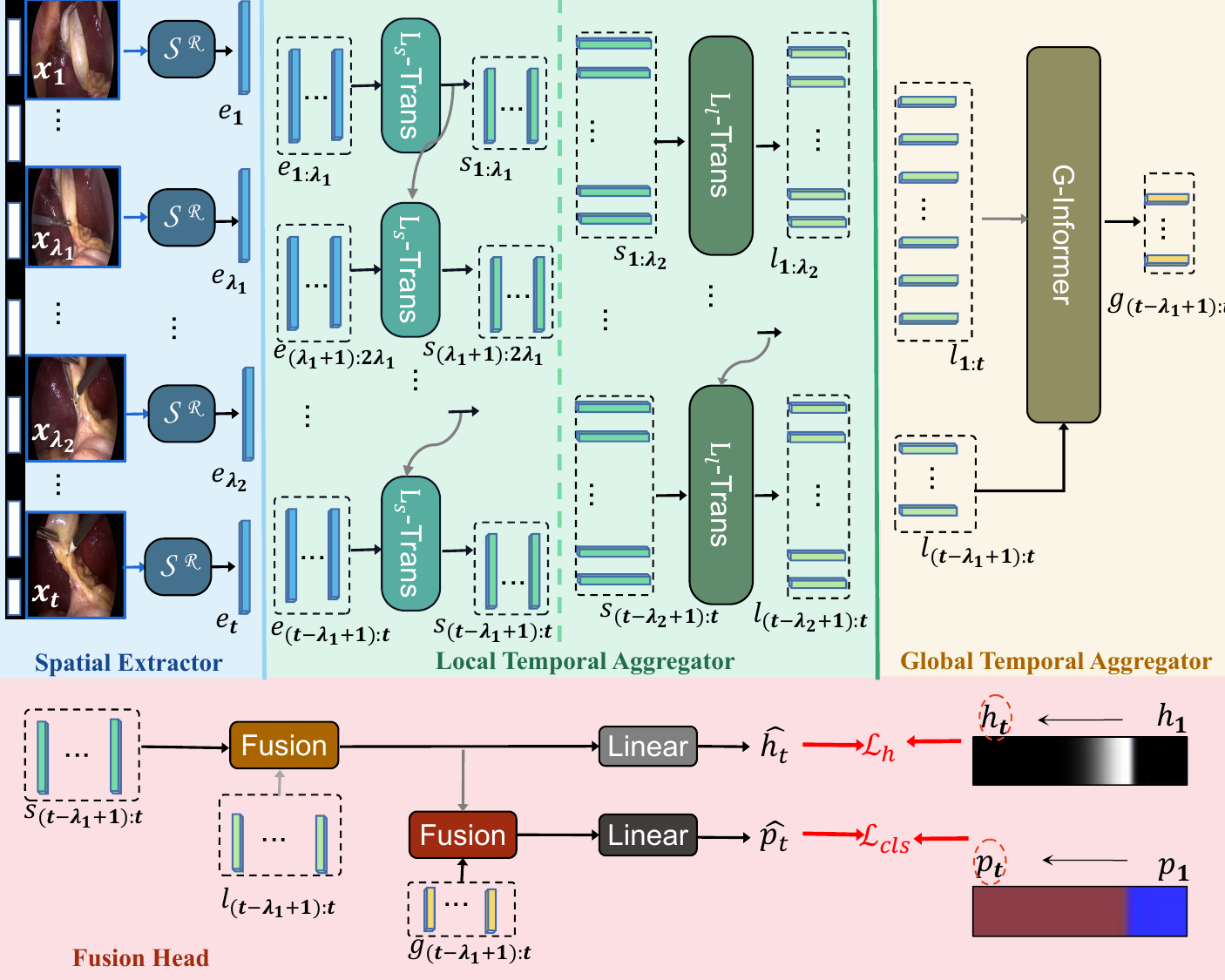}
\caption{The proposed LoViT framework for surgical video phase recognition. The $\mathcal{S}^\mathcal{R}$ module extracts temporally-rich spatial features $e$ from each video frame $x$. Two cascaded L-Trans modules ($\mathrm{L}_s$-Trans and $\mathrm{L}_l$-Trans) output local temporal features $s$ and $l$ with inputs of different local window sizes ($\lambda_1$ and $\lambda_2$). G-Informer captures the global relationships to generate the temporal feature $g$. A fusion head combines the multi-scale features $s$, $l$, and $g$, followed by two linear layers that learn a phase transition map $\hat{h}_t$ and a phase label $\hat{p_t}$ of the current $t$-th video frame $x_t$. Modules with the same color share the same weight. During training, $\mathcal{S}^\mathcal{R}$ is trained separately and its weights are then frozen to train the other temporal modules of LoViT.}
\label{fig:lovit}
\end{center}
\end{figure*}
\subsection{Overview of LoViT architecture}
%\noindent{\textbf{Overview of LoViT architecture.}}
Our LoViT embodies a temporally-rich spatial feature extractor followed by a multi-scale temporal feature aggregator. Specifically, the temporally-rich spatial feature extractor embeds surgical video frames, and then feeds them to the multi-scale temporal feature aggregator, where includes Transformer-based module L-Trans for abstracting short fine-grained information, such as actions and tools, and Informer-based module G-Informer for processing long-term information, such as key clip information of current phase and the relationship among phases. Moreover, a multi-scale temporal fusion head integrates local and global features together that is used for classifying surgical phases with the support of phase transition-aware supervision.

\subsection{Temporally-rich spatial feature extractor}
%\noindent{\textbf{Temporally-rich spatial feature extractor.}} 
\begin{figure}
\centering
\includegraphics[width=\linewidth]{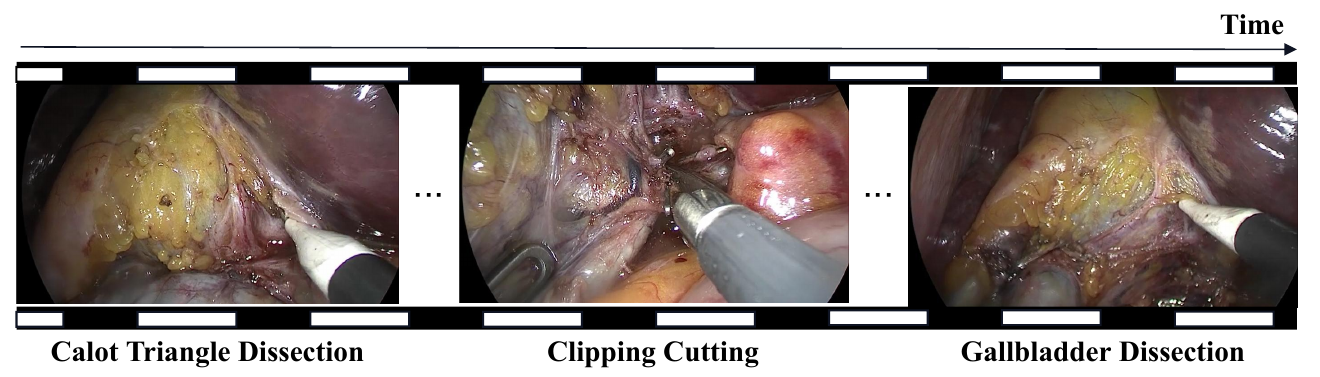}
\caption{Example of similar frames (first and third) corresponding to different phases in Cholec80 dataset~\cite{DBLP:journals/tmi/TwinandaSMMMP17}.}
\label{fig:confusionfig}
\centering
\end{figure}
% TODO I'm not sure you need all this text here which I assume appears already in the previous sections when introducing the temporally-rich spatial feature extractor. From here <-------
Currently, surgical phase recognition methods typically rely on long sequences of frame inputs due to the strong interdependence of surgical phases. However, the duration of surgical videos often lasts for several hours, making it difficult to train a model in an end-to-end manner. To overcome this challenge, a two-step model is widely used, where a spatial feature extractor is trained first to encode spatial features, followed by training a temporal feature extractor with the encoded features.
The prevalent approach to training the spatial feature extractor involves using frame-level supervision, where each image is input and the expected output is the corresponding phase. However, similar actions and screens appearing in different phases during a surgical video can result in confusion when recognizing the phases, as depicted in Fig.~\ref{fig:confusionfig}. This can lead to insufficient training of the spatial feature extractor if only relying on frame-level supervision. %since confusion arises when training a spatial feature extractor module, a problem that can result in over-fitting. 
Given a video frame, its corresponding phase relies on current frame itself and previous frames' features. So only current video frame is insufficient to recognize a phase label. 
% TODO to here ----->
\begin{figure}
\includegraphics[width=1.0\linewidth]{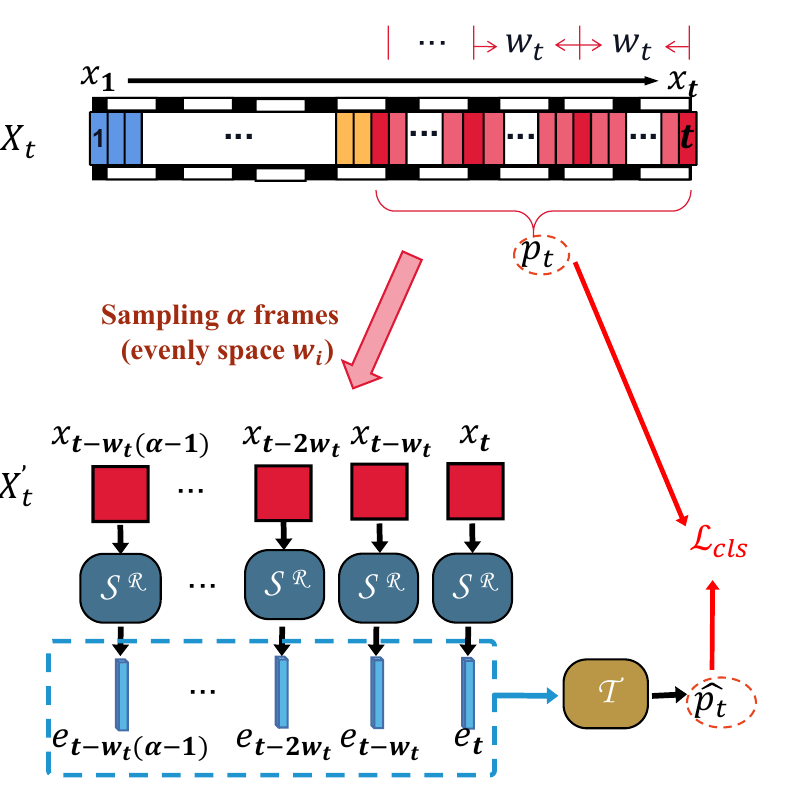}
\caption{The architecture of training the temporally-rich spatial feature extractor. During the $t$-th frame training, a video stream $X_t = \{x_j\}_{j=1}^t$ is sampled at evenly spaced intervals $w_t$ from the start of the current phase to the current frame, producing $X_t' \subseteq X_t$. Each frame $x \in X_t'$ is embedded using the spatial feature extractor $\mathcal{S}^\mathcal{R}$, then grouped into a feature sequence (with a blue dashed box). A temporal aggregator $\mathcal{T}$ follows to add temporal information for recognition. The predicted phase $\hat{p_t}$ is compared to the corresponding ground truth phase $p_t$ to compute a cross-entropy loss. We will throw $\mathcal{T}$ and only retain $\mathcal{S}^\mathcal{R}$ for spatial feature extraction after the training stage.}
\label{fig:spatial_training_visulize}
\end{figure}
We propose a new method to train a temporally-rich spatial feature extractor as shown in Fig.~\ref{fig:spatial_training_visulize}, which builds a more accurate ~(image set $\rightarrow$ phase) mapping relationship instead of (image $\rightarrow$ phase) mapping relationship.

We start by selecting a set of frames $X_t'$ from image sequence $X_t$ and embedding each image frame $x \in X_t'$ into a spatial feature vector $e \in \mathbb{R}^{D_s}$ using the spatial feature extractor $\mathcal{S}^\mathcal{R}$. $D_s$ is the dimension of the spatial feature. These features are then grouped into a feature sequence $E_t$, which is inputted into the temporal aggregator $\mathcal{T}$ for surgical phase classification. $\mathcal{S}^\mathcal{R}$ and $\mathcal{T}$ are jointly trained in an end-to-end manner, leading to a temporally-rich spatial feature extractor $\mathcal{S}^\mathcal{R}$.
Due to limitations in computing resources and the assumption that key temporal information typically appears at the beginning~($b_p$-th frame in the video) of each phase $p$, we only select a small number of $\alpha = 30$ image frames from the beginning $x_{b_{p_t}}$ of the current phase $p_t$ 
% TODO it should be lower-case X_i otherwise it would refer to the entire video
up to the current frame $x_t$ at equal intervals $w_t = \lceil \frac{t - b_{p_t}}{\alpha} \rceil$ to form the image set $X_t'$:
% Fi should be defined earlier, even before T
\begin{equation}
\setlength{\belowdisplayskip}{3pt}
X_t'=(x_{t-w_t(\alpha-1)}, x_{t-w_t(\alpha-2)}, ..., x_t) \label{eq:feature_extractor}.
\end{equation}
 Then the predicted phase $\hat{p_t}$ of $t$-th frame is formulated as:
\begin{equation}
\begin{aligned}
\setlength{\abovedisplayskip}{3pt}
% TODO Use E_i rather than introducing a new variable R_i
\hat{p_t} &= \mathcal{T}(E_t)\\
%&=\mathcal{T}(\mathcal{S}_\mathcal{R}(\mathcal{X}_i))\\
&=\mathcal{T}((\mathcal{S}^\mathcal{R}(x_{t-w_t(\alpha-1)}), \mathcal{S}^\mathcal{R}(x_{t-w_t(\alpha-2)}), ..., \mathcal{S}^\mathcal{R}(x_t)))
\label{eq:tempreallabel}.
\end{aligned}
\end{equation}
% It might be good to replace ViT by S in the diagram
% could we replace theta by a variable, for instance 'l'

% Please mention what loss function you use
To this end, inspired by the Transformer-based model AVT~\cite{DBLP:conf/iccv/GirdharG21} that performs efficiently in action anticipation tasks of short videos, we use the proposed architecture of AVT, whereby spatial feature extractor $\mathcal{S}^\mathcal{R}$ is based on ViT~\cite{DBLP:conf/iclr/DosovitskiyB0WZ21} and temporal aggregator $\mathcal{T}$ is based on a Transformer with causal-masked attention. Moreover, we use the cross-entropy loss for training this model.
Note that we throw out the temporal aggregator $\mathcal{T}$ after training the spatial feature extractor $\mathcal{S}^\mathcal{R}$, and use the fixed weight of $\mathcal{S}^\mathcal{R}$ to embed each image frame $x$ into spatial feature $e$, which is fed to the following temporal feature aggregator.

% TODO REMOVE 4.1: removing for now for consistency
%\subsection{Multi-scale temporal feature aggregator}

\subsubsection{Local temporal feature aggregator}

    \begin{figure}
\centering
\includegraphics[width=0.8\linewidth]{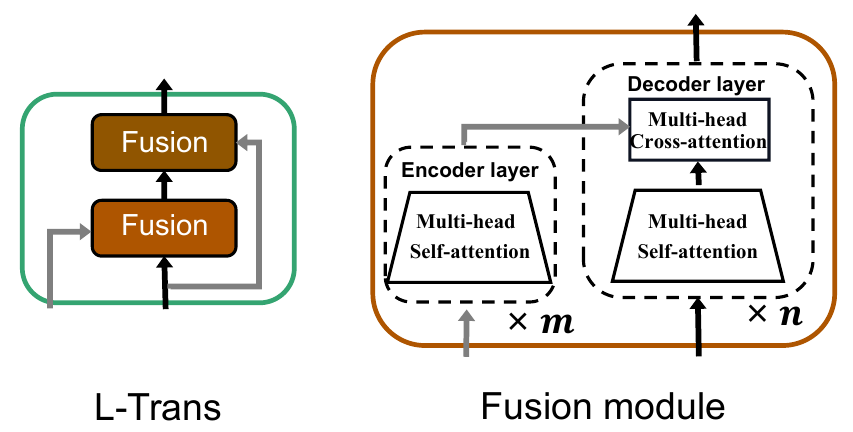}
\caption{\textbf{L-Trans:}~The L-Trans adopts two cascaded fusion modules to  process two-branch temporal inputs~(grey line and black line). \textbf{Fusion module:}~It consists of an encoder and a decoder. The encoder is comprised of an $m$-layer self-attention layer for the grey line input, and the decoder is composed of an $n$-layer cascaded self-attention with cross-attention for processing the encoder's output and the black line input.}
\label{fig:localtrans_and_fusionmodule}
\centering
\end{figure}
%\noindent{\textbf{Local temporal feature aggregator.}}
We designed a local Transformer-based temporal feature aggregator~(\textbf{L-Trans}) to extract local fine-grained temporal information, which is visualized in Fig.~\ref{fig:localtrans_and_fusionmodule}. 
L-Trans begins by analyzing the current local temporal features with the assistance of the previous clip's output, using a fusion module. A second fusion module then refines the output by incorporating initial current features. Fig.~\ref{fig:localtrans_and_fusionmodule} also depicts the fusion module, which consists of an $m$-layer self-attention encoder for auxiliary features and an $n$-layer cascaded self-attention module with cross-attention for the encoder's output and the decoder branch input.

The self-attention mechanism reduces the length between every network signal to the shortest $\mathcal{O}(1)$ through the dot-product computation between every two signals and avoids the recurrent structure, whereby Transformer is used for temporal feature aggregation. 
 The self-attention mechanism, per \cite{DBLP:conf/nips/VaswaniSPUJGKP17}, is formulated as:
\begin{equation}
\setlength{\belowdisplayskip}{3pt}
\mathcal{A}(Q, K, V) = {\rm Softmax}(\frac{QK^T}{\sqrt{d_k}})V,
\label{eq:selfattention}
\end{equation}
where $Q \in \mathbb{R}^{L_Q\times d}$, $K \in \mathbb{R}^{L_K\times d}$, $V\in \mathbb{R}^{L_V\times d}$, and $d$ is the input dimension. $Q$, $K$, and $V$ represent the standard matrices referred to as query, key, and value in Transformer-based architectures, respectively. 
In contrast, cross-attention refers to the attention mechanism that takes into account both the query and key inputs from different sources. 

% TODO I think you can delete this paragraph below since it repeats what is mentioned in the one after

% Could you make sure these notation of thetas is reflected in Fig 1.
% Also could you donote these as theta_s and theta_l to refer to the Ls-Trans and Ll-Trans?
To capture local temporal information of different granularities, we utilize two cascaded L-Trans: a small one ($\mathbf{L}_s$\textbf{-Trans}) and a large one ($\mathbf{L}_l$\textbf{-Trans}), each receiving input sequences of length $\lambda_1$ and $\lambda_2$, respectively. 
We first feed the spatial feature set $e$ resulting from our temporally-rich spatial feature extractor to ${\rm{L}}_s$-Trans and output the small-local feature $s$. Note that we also include the previous output of  ${\rm{L}}_s$-Trans as input. 
Then, we feed the sequence $s$ to ${\rm{L}}_l$-Trans to obtain the large-local feature $l$. Similarly, note that we also include the previous output of ${\rm{L}}_l$-Trans as input.
To speed up the training time and reduce the training memory, we only calculate the gradient of the last clip $(x_{t-\lambda+1}, ..., x_{t})$ and drop all the previous gradients when recognising $t$-th frame for training.

\subsubsection{Global temporal feature aggregator}
%\noindent{\textbf{Global temporal feature aggregator.}}
% Same, please improve captions of figure
%\input{Images/Fig_Global_Trans}
%Surgical videos are characterised by a strong time dependency between phases, variability across same type of intervention, and with actions that might appear similar across phases.
%Due to the high stability of surgical procedure, there is a strong time dependency between phases. 
%The model could recognise current phase through such relationships even the same local action would appear in different phases. 
%Key information is generally found at the beginning of every phase, which is crucial for recognising the phase of a given frame. 
%Since surgical videos are typically long, phase recognition requires models with the ability to process long sequences of video frames and extract their temporal relationships. 
%Both of them belong to the long-term relationship, and require our model have the ability to process long sequence input. 
%We propose to use a more efficient implementation of a Transformer-based architecture to process long sequences of information from surgical videos. 
Despite the self-attention mechanism inherent in Transformer-based architectures has the capability of extracting temporal relationships (see Eq.~\ref{eq:selfattention}), it requires quadratic dot-product $\mathcal{O}(\textit{L}_Q \textit{L}_K)$ time computation and memory usage, which limits the processing of long sequences.
To overcome this limitation, we propose the use of Informer~\cite{DBLP:conf/aaai/ZhouZPZLXZ21}, which consists of a more efficient implementation of the self-attention mechanism called \textit{ProbSparse}, whereby only a few dot-product pairs contribute to the major attention. \textit{ProbSparse} reduces time and memory usage to $\mathcal{O}(\textit{L} \ln(\textit{L}))$ where $L$ represents the input sequence length.
\textit{ProbSparse} is formulated as:

\begin{equation}
\setlength{\abovedisplayskip}{3pt}
\setlength{\belowdisplayskip}{3pt}
\mathcal{A}_{PS}(Q, K, V) = {\rm Softmax}(\frac{\overline{Q}K^T}{\sqrt{d_k}})V,
\label{eq:sparseselfattention}
\end{equation}
% TODO same saze as Q?
where $q_i$ represents the $i$-th row in Q, and $\overline{Q}$ is a sparse matrix of the same size as $q$, containing only the top $u$ queries under the sparsity max-mean measurement:

\begin{equation}
\setlength{\abovedisplayskip}{3pt}
\setlength{\belowdisplayskip}{3pt}
\overline{M}(q_i, K) = \underset{j}{\max}(\frac{q_ik_j^T}{\sqrt{d_k}}) - \frac{1}{L_K}\sum_{j=1}^{L_K}\frac{q_ik_j^T}{\sqrt{d_k}}.
\label{eq:measurementsparseselfattention}
\end{equation}

Under the long tail distribution, we randomly sample $U = L_K\ln(L_Q)$ dot-product pairs to calculate $\overline{M}(q_i, K)$, and fill the rest with zeros.

We designed a Global Temporal Informer (\textbf{G-Informer}) to capture long-range dependencies more efficiently. The G-Informer framework contains two branches for processing two types of inputs: a long local feature sequence $(l_1,..., l_t)$ and the current local feature $(l_{t-\lambda_1+1},...,l_t)$. Unlike the fusion model used for L-Trans, the first input branch of G-Informer for the long sequence uses the \textit{ProbSparse} self-attention mechanism.

% We need more details here, it appears to basic, we can support this with another equation to clarify the order of the inputs during fusion 
%\subsubsection{Multi-scale temporal feature fusion head}
\subsection{Multi-scale temporal feature fusion head.}
%\input{Images/Fig_Fuse_Trans}
%LoViT uses L-Trans and G-Informer to aggregate local and global temporal features, respectively via a multi-scale temporal feature fusion head. 
% Next sentence is unclear
Feature down-sampling and sparse attention unfortunately lead to the loss of fine-grained characteristics in G-Informer. To overcome this limitation, we employ a multi-scale temporal feature fusion head to combine the local (small and large) and global features from L-Trans and G-Informer, respectively. 
As illustrated at the bottom of Fig.~\ref{fig:lovit}, the multi-scale head contains two fusion modules. 
The first one is to merge the short and the long local temporal features, $(s_{t-\lambda_1+1},...,s_t)$ and $(l_{t-\lambda_1+1},...,l_t)$, obtained from ${\rm{L}}_s$-Trans and ${\rm{L}}_l$-Trans, respectively. Subsequently, another fusion module is utilized to merge the fused local features with the global temporal features $(g_{t-\lambda_1+1},...,g_t)$ obtained from G-Informer.

%\subsection{Heatmap-guided supervision}
\subsection{Phase transition-aware supervision.}

\begin{figure}
\centering
\includegraphics[width=0.8\linewidth]{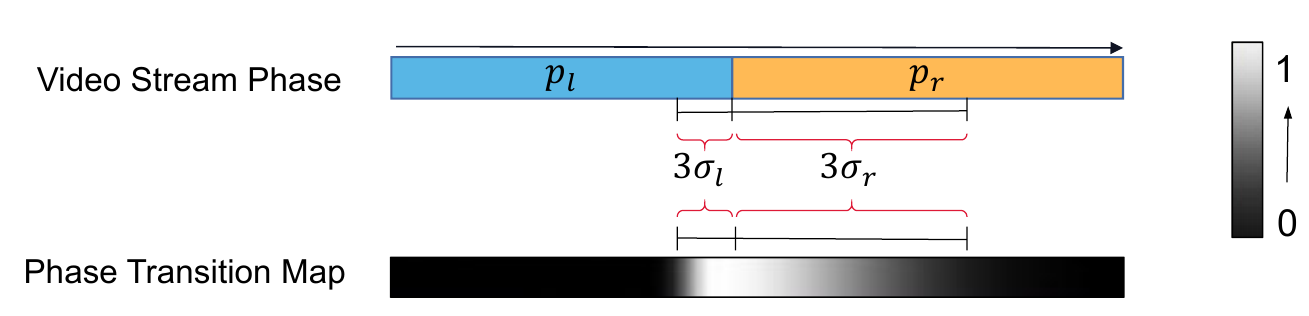}
\caption{The example of building phase transition map. We project phase transition area onto a phase transition map using a left-right asymmetric Gaussian kernel where left- and right-side kernel lengths are $3\sigma_l$ and $3\sigma_r$ respectively. $p_l$ and $p_r$ mean adjacent different phases.}
\label{fig:heatmap}
\centering
\end{figure}
To be able to give importance to previous key information, especially during phase transitions, we project transition frames, which are adjacent to other phase frames, onto a phase transition map $h$ using a one-dimensional asymmetric Gaussian kernel, as shown in Fig.~\ref{fig:heatmap}. The map value of the $t$-th frame is formulated as:

\begin{small}
\begin{equation}
\setlength{\abovedisplayskip}{3pt}
\setlength{\belowdisplayskip}{3pt}
h_t = 
\left \{
    \begin{matrix}
    \exp (-\frac{(t - b_{p_t})^2}{2\sigma^2_l}),\ &b_{p_t} - 3\sigma_l \textless t \textless b_{p_t} \\
    \exp (-\frac{(t - b_{p_t})^2}{2\sigma^2_r}),\ &b_{p_t} \textless t \textless b_{p_t} + 3\sigma_r\\
    0, &\ otherwise
    \end{matrix}
\right.
    \label{eq:heatmap}
\end{equation}
\end{small}where the left-side kernel length of the transition map is $3\sigma_l$ and the right-side kernel length is $3\sigma_r$, both departing from $b_{p_t}$, the position of the frame where the phase transition occurs.
Accordingly, the loss function of our proposed LoViT model is a weighted sum of the phase transition map loss and phase class loss:

% Chat now same variable as in the beginning, I don't think we need that one in Eq. 1
\begin{equation}
\setlength{\abovedisplayskip}{3pt}
\setlength{\belowdisplayskip}{3pt}
\mathcal{L}^* = \mathcal{L}_1(\hat{h}) + \mathcal{L}_{CE}(\hat{p}),
\label{eq:lossfuction}
\end{equation}
% there might be an error here, since the inputs to loss functions are different, please refer to H_hat and p_hat as in the Eq.
where $\mathcal{L}_1(\hat{h})$ refers to the $L_1$ loss between predicted phase transition map value $\hat{h}$ and its ground truth $h$, and $\mathcal{L}_{CE}(\hat{p})$ refers to cross-entropy loss between predicted phase $\hat{p}$ and its ground truth $p$.
%Note we only keep spacial feature extractor $S$ and will drop temporal aggregator $T$ while training the LoViT and during test inference.

\section{Experimental Design}

\subsection{Datasets.}
We extensively performed experiments on two publicly available surgical video datasets, namely Cholec80~\cite{DBLP:journals/tmi/TwinandaSMMMP17} and AutoLaparo~\cite{MICCAI2022_data}, capturing cholecystectomy and hysterectomy surgical interventions, respectively. 
%These datasets are described below. 
Cholec80~\cite{DBLP:journals/tmi/TwinandaSMMMP17} consists of 80 high-resolution, of either 1920$\times$1080 or 854$\times$480 pixels, laparoscopic surgical videos with an average video duration of 39 minutes at 25 frames-per-second (fps). These databases are provided with manual annotations done by surgeons indicating the surgical phase each video frame belongs to and the tools appearing in the scene. 
Cholec80 videos consist of seven phases.
For this study, we only use phase annotations. 
For a fair comparison with previous methods~\cite{DBLP:journals/tmi/JinD0YQFH18, DBLP:journals/tmi/TwinandaSMMMP17, DBLP:conf/miccai/YiJ19, DBLP:conf/miccai/GaoJLDH21}, we keep intact the splitting of the dataset into first 40 videos for training and the remaining 40 videos for testing. AutoLaparo~\cite{MICCAI2022_data} consists of 21 videos with 7 phases, recorded at 25~Hz of a resolution of 1920$\times$1080 pixels with an average video duration of 66 minutes. We split the dataset into 10 videos for training, 4 videos for validation and 7 videos for testing following~\cite{MICCAI2022_data}. 
Note that, similar to other works presented in the literature~\cite{DBLP:journals/tmi/JinD0YQFH18, DBLP:journals/tmi/TwinandaSMMMP17, DBLP:conf/miccai/GaoJLDH21}, we sampled both datasets into 1~fps and resized the frame size to 250$\times$250 pixels.

\subsection{Implementation details.}
Our method is implemented with the PyTorch framework~\cite{DBLP:conf/nips/PaszkeGMLBCKLGA19}. All experiments were carried out on an Intel Xeon W-2195 CPU~(2.3 GHz), 125GB RAM, and a single NVIDIA Tesla V100 GPU. Following the AVT model~\cite{DBLP:conf/iccv/GirdharG21}, our temporally-rich spatial feature extractor is a 12-head, 12-layer Transformer encoder model that uses the ViT-B/16 architecture, which is pretrained on ImageNet-1K~(IN1k)~\cite{Imagenet_09} with input image size of 248$\times$248 pixels and output size of $D_s = 768$D representations.
We trained the feature extractor with SGD+momentum for 35 epochs, with a 5-epoch warmup~~\cite{DBLP:journals/corr/GoyalDGNWKTJH17} and cosine annealed decay. The rest of LoViT was trained for 50 epochs with SGD+momentum, weight decay of $1e-5$, learning rate of $3e-4$, and a 5-epoch warmup and 45-epoch cosine annealed decay.
During experiments, LoViT was fed 3000 video frames and produced outputs with dimensions of 512, 64, and 8 for ${\rm{L}}_s$-Trans, ${\rm{L}}_l$-Trans, and G-Informer, respectively. Both ${\rm{L}}_s$-Trans and ${\rm{L}}_l$-Trans have a fusion module each with a 2-layer encoder and a 2-layer decoder. G-Informer has a 2-layer encoder and a 1-layer decoder. The fusion modules in the multi-scale head have a 2-layer encoder and a 1-layer decoder. To improve efficiency, the gradients for ${\rm{L}}_s$-Trans were calculated using the last $\lambda_1 = 100$ frames and the gradients for ${\rm{L}}_l$-Trans were calculated using the last $\lambda_2 = 500$ frames. About the proposed phase transition map, we set $\sigma_l = 3$ and $\sigma_r = 12$.

\subsection{Evaluation.}
In this paper, we investigate the performance of LoViT in comparison with state-of-the-art approaches, followed by extensive ablation experiments to demonstrate the effect its different components have on a surgical phase recognition task.
In accordance with previous work, we use four frequently-used measures in surgical phase recognition, namely accuracy~(AC), precision~(PR), recall~(RE), and Jaccard~(JA). 
AC refers to the percentage of correctly recognized frames and is video-based. 
However, the video class is imbalanced, and the short phases only have little impact on the whole video's accuracy. To evaluate our model in multiple dimensions, we further adapt class-~(phase-) level precision, recall, and Jaccard, which represent positive predictive value, positive rate, and intersection rate of recognition versus ground truth, respectively.

\section{Results}

\subsection{Comparison with state-of-the-art methods}

\begin{table}[]
\begin{centering}
%\captionsetup{justification=centering}
\caption{The results~(\%) of different state-of-the-art methods on both the Cholec80 and AutoLaparo datasets. The best results are marked in bold. Note that the `*' denotes methods based on multi-task learning that requires extra tool labels, and '+' denotes the use of 10-second relaxed boundary metrics\protect\footnotemark[1].}
\resizebox{\linewidth}{!}{
\renewcommand\arraystretch{1.2}
\begin{tabular}{cccccc}
\hline    
\multirow{2}{*}{Dataset} &\multirow{2}{*}{Method} &Video-level Metric  &\multicolumn{3}{c}{Phase-level Metric}\\\cmidrule{3-6} 
 & & $\mathrm{Accuracy}\uparrow$ &$\mathrm{Precision}\uparrow$ &$\mathrm{Recall}\uparrow$ &$\mathrm{Jaccard}\uparrow$\\\hline

\multirow{9}{*}{Cholec80}

&EndoNet~\cite{DBLP:journals/tmi/TwinandaSMMMP17}$^{*+}$ &$81.7 \pm 4.2$ &$73.7 \pm 16.1$ &$79.6 \pm 7.9$       &-    \\
&MTRCNet-CL~\cite{DBLP:journals/mia/JinLDCQFH20}$^{*+}$ &$89.2 \pm 7.6$ &$86.9 \pm 4.3$ &$88.0 \pm 6.9$ 
&-\\
\cline{2-6}
&PhaseNet~\cite{DBLP:journals/corr/TwinandaMMMP16}$^{+}$
&$78.8\pm 4.7$        &$71.3 \pm 15.6$      &$76.6 \pm 16.6$       & -    \\
&SV-RCNet~\cite{DBLP:journals/tmi/JinD0YQFH18}$^{+}$ &$85.3 \pm 7.3$    &$80.7 \pm 7.0$ &$83.5 \pm 7.5$
& -\\
&OHFM~\cite{DBLP:conf/miccai/YiJ19}$^{+}$ & $87.3 \pm 5.7$     & -       & -        & $67.0 \pm 13.3$\\
&TeCNO~\cite{DBLP:conf/miccai/CzempielPKSFKN20}$^{+}$ & $88.6 \pm 7.8$      & $86.5 \pm 7.0$       & $87.6 \pm 6.7$        &  $75.1 \pm 6.9$\\

&TMRNet~\cite{DBLP:journals/tmi/JinLCZDH21}$^{+}$ &$90.1 \pm 7.6$  &$90.3 \pm 3.3$  &$89.5 \pm 5.0$  &$79.1 \pm 5.7$ \\

&Trans-SVNet~\cite{DBLP:conf/miccai/GaoJLDH21}$^{+}$ & $90.3 \pm 7.1$      & $ \bf{90.7 \pm 5.0}$       & $88.8 \pm 7.4$        & $ 79.3 \pm 6.6$\\
&LoViT~(ours)$^{+}$ & $\bf{92.40 \pm 6.3}$      & $89.9 \pm 6.1$       & $ \bf{90.6 \pm 4.4}$        & $\bf{81.2 \pm  9.1}$\\
\cline{2-6}
&Trans-SVNet & $89.1 \pm 7.0$      & $\bf{84.7}$       & $83.6$        & $72.5$\\
&AVT~\cite{DBLP:conf/iccv/GirdharG21} & $86.7 \pm 7.6$      & $77.3$       & $82.1$        & $66.4$\\
&LoViT~(ours)~ & $\bf{91.5 \pm 6.1}$      & $83.1$       & $\bf{86.5}$        & $\bf{74.2}$\\
\bottomrule[1pt] 

\multirow{6}{*}{AutoLaparo} &SV-RCNet   &$75.6$  &$64.0$  &$59.7$  &$47.2$ \\
&TMRNet &$78.2$  &$66.0$  &$61.5$  &$49.6$ \\
&TeCNO  &$77.3$  &$66.9$  &$64.6$  &$50.7$ \\
&Trans-SVNet   & $78.3$  &$64.2$  &$62.1$  &$50.7$  \\
&AVT~ & ${77.8 \pm 9.4}$      & ${68.0}$       & ${62.2}$        & ${50.7}$\\
&LoViT~(ours)~ & $\bf{81.4 \pm 7.6}$      & $\bf{85.1}$       & $\bf{65.9}$        & $\bf{55.9}$\\
\bottomrule[1pt]
\end{tabular}}

\label{tab:sota}
\end{centering}
\end{table}
\footnotetext[1]{Refer to the code: \url{https://github.com/YuemingJin/TMRNet/blob/main/code/eval/result/matlab-eval/Evaluate.m}. These metrics consider predictions that fall into neighbouring phases as correct, within a 10-second window around the phase transition. We do not support this approach of allowing boundary mistakes to improve metric values, as phase transition prediction is also an important indicator of a model's ability. However, we include this approach to ensure a fair comparison with previous methods.}
To evaluate the effectiveness of our proposed method, we conducted a comparative analysis of LoViT against other state-of-the-art approaches that are relevant to action anticipation and surgical phase recognition tasks. This analysis was carried out on two datasets: Cholec80~\cite{DBLP:journals/tmi/TwinandaSMMMP17} and AutoLaparo~\cite{MICCAI2022_data}.

The quantitative comparison on Cholec80 dataset is organised in the top section of Table~\ref{tab:sota}. 
Note that we re-implemented Trans-SVNet using the model weights provided with the original manuscript. For AVT, we used 30 input frames based on the publicly available code provided with their published manuscript.
The results of the other state-of-the-art methods were extracted verbatim from their respective published works.
We did not include other state-of-the-art methods such as OperA~\cite{DBLP:conf/miccai/CzempielPOKBN21} in our study as they use a different dataset split for training and testing and do not have publicly available code accompanying their manuscript.
Table~\ref{tab:sota} demonstrates that LoViT outperforms other methods on most metrics, with the exception of precision and recall on the Cholec80 dataset. Specifically, for precision, LoViT ranks third after MTRCNet-CL and Trans-SVNet, and for recall, it ranks second after MTRCNet-CL. LoViT achieves higher accuracy than the current state-of-the-art benchmark Trans-SVNet, with a margin of 2.4 pp (percentage points). 
Even compared to MTRCNet-CL, a multi-task learning method that requires additional information in the form of tool labels, LoViT exhibited higher performance with a margin of 2.3 pp accuracy. LoViT also outperformed AVT, which is the champion model for action anticipation and is similar to phase recognition, by 4.8 pp accuracy.
Furthermore, our model showed a lower standard deviation of the accuracy of approximately 1.5 pp compared to Trans-SVNet, MTRCNet-CL, and AVT.

The quantitative comparison on AutoLaparo dataset is depicted in the bottom section of Table~\ref{tab:sota}. 
We reference~\cite{MICCAI2022_data} to populate Table~\ref{tab:sota} with existing methods evaluated on this dataset. 
We observe that TMRNet~\cite{DBLP:journals/tmi/JinLCZDH21}, TeCNO~\cite{DBLP:conf/miccai/CzempielPKSFKN20}, and Trans-SVNet~\cite{DBLP:conf/miccai/GaoJLDH21} perform similarly on this dataset with average accuracy of 77\%.
% changeable? do you mean challenging?
% TODO please add reference of "Evidence suggest ..."
Evidence~\cite{MICCAI2022_data} suggests that AutoLaparo is more challenging because of the complex workflow with a small dataset scale. 
Compared to the state-of-the-art method Trans-SVNet, we observed LoViT had a higher performance with a 3.1 pp accuracy margin. 
Besides video-level accuracy, we highlight that reported phase-level metrics are crucial due to the imbalance of phase distribution. 
Compared with Trans-SVNet, an increase of 20.9 pp~(64.2\% $\rightarrow$ 85.1\%), 3.8 pp~(62.1\% $\rightarrow$ 65.9\%), and 5.2 pp~(50.7\% $\rightarrow$ 55.9\%) margins were achieved using LoViT in relation to precision, recall, and Jaccard, respectively. 
In summary, LoViT consistently outperformed in both video-level and phase-level as evidenced by these evaluation metric results.

\begin{figure}[ht]
\centering
\includegraphics[width=\linewidth]{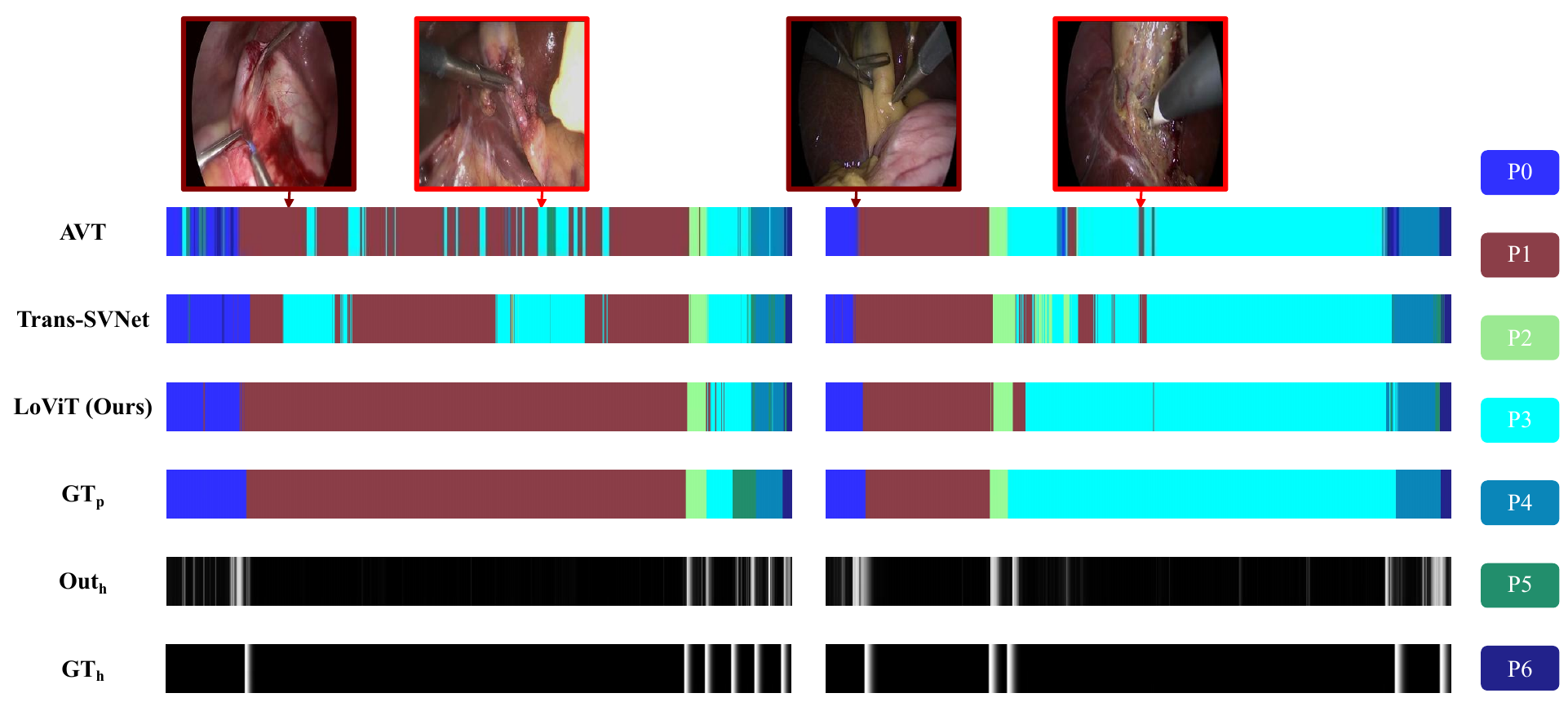}
\caption{Qualitative comparisons with some other methods on Cholec80 dataset. The first line presents some images in the video corresponding to the moment pointed by the red arrow, where light red present wrong examples of both AVT and Trans-SVNet, and dark red present wrong examples of only Trans-SVNet. The following four lines represent the phase results recognised by different methods and corresponding Ground Truth $\mathrm{GT_p}$. The last two lines mean the output of heatmap by proposed LoViT $\hat{h}$ and its Ground Truth $\mathrm{GT_h}$}.
\label{fig:visualize_comparewithsota}
\centering
\end{figure}

To illustrate the performance of our approach in comparison with the state-of-the-art, in Fig.~\ref{fig:visualize_comparewithsota} we present a qualitative comparison of two examples drawn from the Cholec80 testing dataset. 
As observed in Fig.~\ref{fig:visualize_comparewithsota}, for some ambiguous frames shown (first row), Trans-SVNet was unable to classify the correct phase effectively. 
Even though the surgical phases of laparoscopic cholecystectomy are linearly executed, video frames are misclassified  into phases that are strictly nonlinear.
In contrast, LoViT learned a better long-term temporal context than Trans-SVNet, even when a few misclassifications are still nonlinear. Even for some examples~(first row with dark red box), Trans-SVNet performs worse than AVT, which inputs short video clips. This further proves that Trans-SVNet loses some fine-grained and continuity information while processing long videos.
When investigating the performance in learning the heatmap capturing phase transitions, we observed that LoViT's performance is highly accurate compared to the ground truth, as shown in the last two rows of Fig.~\ref{fig:visualize_comparewithsota}.
In our ablation studies, we demonstrate the benefit of including heatmap information since it further helps to extract relationships among phases.

\begin{table}[]
\caption{The results~(\%) of different parts of proposed LoViT on both the Cholec80 and the AutoLaparo datasets. The best results are marked in bold.}
\renewcommand\arraystretch{1.2}
%\begin{center}
\resizebox{\linewidth}{!}{
\begin{tabular}{cccccc}
\hline    
\multirow{2}{*}{Dataset} &\multirow{2}{*}{Model} &Video-level Metric  &\multicolumn{3}{c}{Phase-level Metric}\\\cline{3-6}
 & & $\mathrm{Accuracy}\uparrow$ &$\mathrm{Precision}\uparrow$ &$\mathrm{Recall}\uparrow$ &$\mathrm{Jaccard}\uparrow$\\\hline
%\multicolumn{2}{c|}{VI} & \multirow{2}{*}{Time (s)} \\ \cline{3-12}
%                                      && ODS         & OIS        & ODS        & OIS        & ODS        & OIS        & ODS         & OIS        & ODS         & OIS &\\\midrule[1pt]
\multirow{3}{*}{Cholec80} &L-Trans           & $90.81 \pm 5.85$     & $82.51$       & $86.48$        & $72.92$\\
&G-Informer           & $\bf{91.52 \pm 5.76}$      & $\bf{83.29}$       & $\bf{86.98}$        & $\bf{74.55}$\\
&LoViT           & $91.50 \pm 6.10$      & $83.07$       & $86.5$        & $74.15$\\
\hline
\multirow{3}{*}{AutoLaparo} &L-Trans           & $80.60 \pm 6.93$      & $69.70$       & $65.08$        & $54.02$\\
&G-Informer           & $79.75 \pm 7.32$     & $70.33$       & $63.64$        & $53.43$\\
&LoViT           & $\bf{81.43 \pm 7.35}$      & $\bf{85.07}$       & $\bf{65.85}$        & $\bf{55.90}$\\
\hline
\end{tabular}}
%\end{center}
\label{tab:function_analysis}
\end{table}
\subsection{Evaluating LoViT architecture performance}
We conducted experiments to measure the contributions of the three modules that comprise our proposed model, LoViT. These modules are: 1) Local Temporal Feature Aggregator (L-Trans), 2) Global Temporal Feature Aggregator (G-Informer), and 3) Multi-scale Temporal Feature Fusion Module (MF-Trans). Specifically, we conducted the following experiments:
\begin{itemize}%
\item[$\bullet$] \textbf{L-Trans}: We evaluated the performance of the time aggregation model when it only contained the L-Trans module for capturing local fine-grained features without the integration of global relationships.
\item[$\bullet$] \textbf{G-Informer}: We evaluated the performance of a model that included time aggregation by the G-Informer module following the L-Trans. The coarse-grained information resulting from G-Informer was then directly fed into the classifier.
\item[$\bullet$] \textbf{LoViT}: We evaluated the performance of our full LoViT model, which includes both the temporal local and global Transformers, followed by a multi-scale temporal fusion head.
\end{itemize}

The quantitative experiment results are shown in Table~\ref{tab:function_analysis}. 
Based on the results of L-Trans and G-Informer, it can be observed that G-Informer performed better on the Cholec80 dataset while L-Trans performed better on the AutoLaparo dataset. This suggests that global relationships are more helpful for recognizing the videos in Cholec80, while fine-grained features are more critical for AutoLaparo. This finding is consistent with the fact that most videos in AutoLaparo contain recurring phases, while the workflow in Cholec80 videos is much more linear. Thus, global relationships are more critical for Cholec80, while fine-grained features are more critical for AutoLaparo, as indicated by the experimental results.
Comparing the performance of LoViT with that of L-Trans and G-Informer, we can see that LoViT performs only slightly worse than G-Informer on Cholec80, but outperforms the other two models when evaluated on AutoLaparo. This suggests that while global relationships are important for Cholec80, fine-grained features are crucial for AutoLaparo. Moreover, it is evident that the feature map $g$ produced by G-Informer is more prone to losing fine-grained information than $l$, which is extracted by L-Trans. However, LoViT overcomes this limitation through the multi-scale fusion of $l$ and $g$, allowing the analysis of every frame from different dimensions and resulting in improved recognition accuracy and greater stability.

% TODO Following Luis' suggestion, it might be good to compare these with the spatial features resulting from a spatial extractor trained purely on Image-Net.  Let's discuss before doing it though
\begin{table}
%\begin{center}

\caption{Effects~(\%) of Temporally-rich spacial feature extractor ($\mathcal{R}$) on Cholec80 and  AutoLaparo datasets. The best results are marked in bold.}
\renewcommand\arraystretch{1.2}

\resizebox{\linewidth}{!}{
\begin{tabular}{cccccc}
\hline    
\multirow{2}{*}{Dataset} &\multirow{2}{*}{$\mathcal{R}$} &Video-level Metric  &\multicolumn{3}{c}{Phase-level Metric}\\\cline{3-6}
 & & $\mathrm{Accuracy}\uparrow$ &$\mathrm{Precision}\uparrow$ &$\mathrm{Recall}\uparrow$ &$\mathrm{Jaccard}\uparrow$\\\hline

\multirow{2}{*}{Cholec80} &           & $90.66 \pm 6.89$      & $80.81$        & $85.21$         & $71.42$\\
\cline{2-6}
&\checkmark           & $\bf{91.50 \pm 6.10}$      & $\bf{83.07}$       & $\bf{86.5}$        & $\bf{74.15}$\\
\hline

%\bottomrule[1pt] 
\multirow{2}{*}{AutoLaparo} & & $79.53 \pm 8.36$      & 79.61       & 64.85        & 53.74\\ \cline{2-6}
&\checkmark           & $\bf{81.43 \pm 7.35}$      & \bf{85.07}       & \bf{65.85}        & \bf{55.90}\\
\hline

%\bottomrule[1pt] 
\end{tabular}}
%\end{center}
\label{tab:sapcialextractor}
\end{table}
\begin{figure}
\centering
\includegraphics[width=0.8\linewidth]{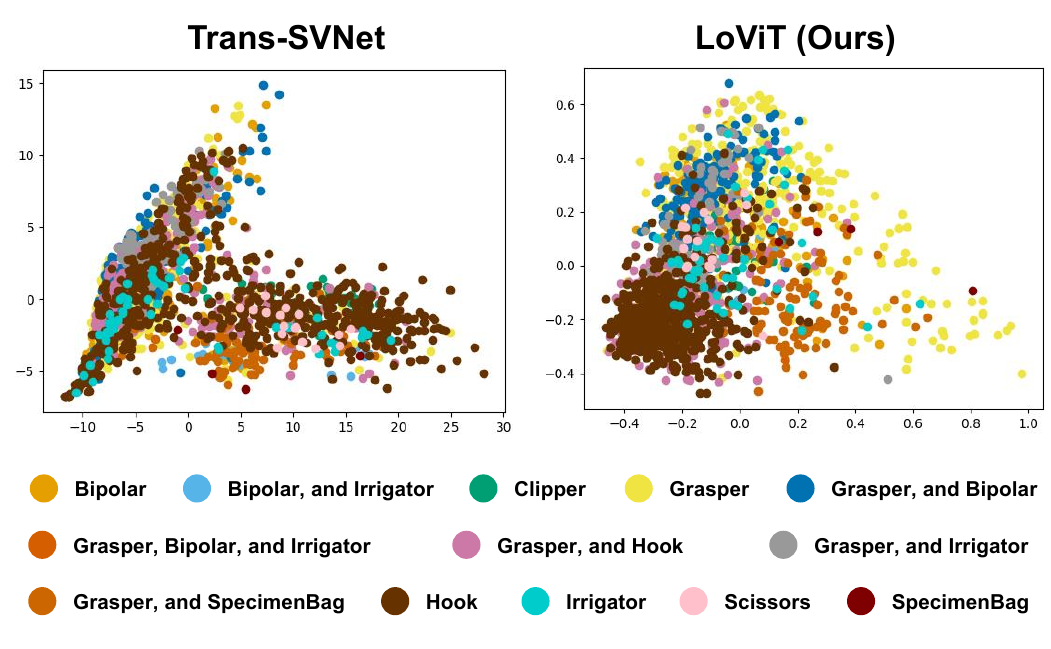}
\caption{Visualization for the spatial feature distribution of different extractors. Point set: Video frames of Video 60 in Cholec80. Different colors: different tool annotations. First column: the spatial feature distribution of the frame-only spatial feature extractor in Trans-SVNet. Second column: the spatial feature distribution of the temporally-rich spatial feature extractor in our LoViT.}
\label{fig:visual_spatial_wholevideo}
\centering
\end{figure}
\begin{figure}
\centering
\includegraphics[width=\linewidth]{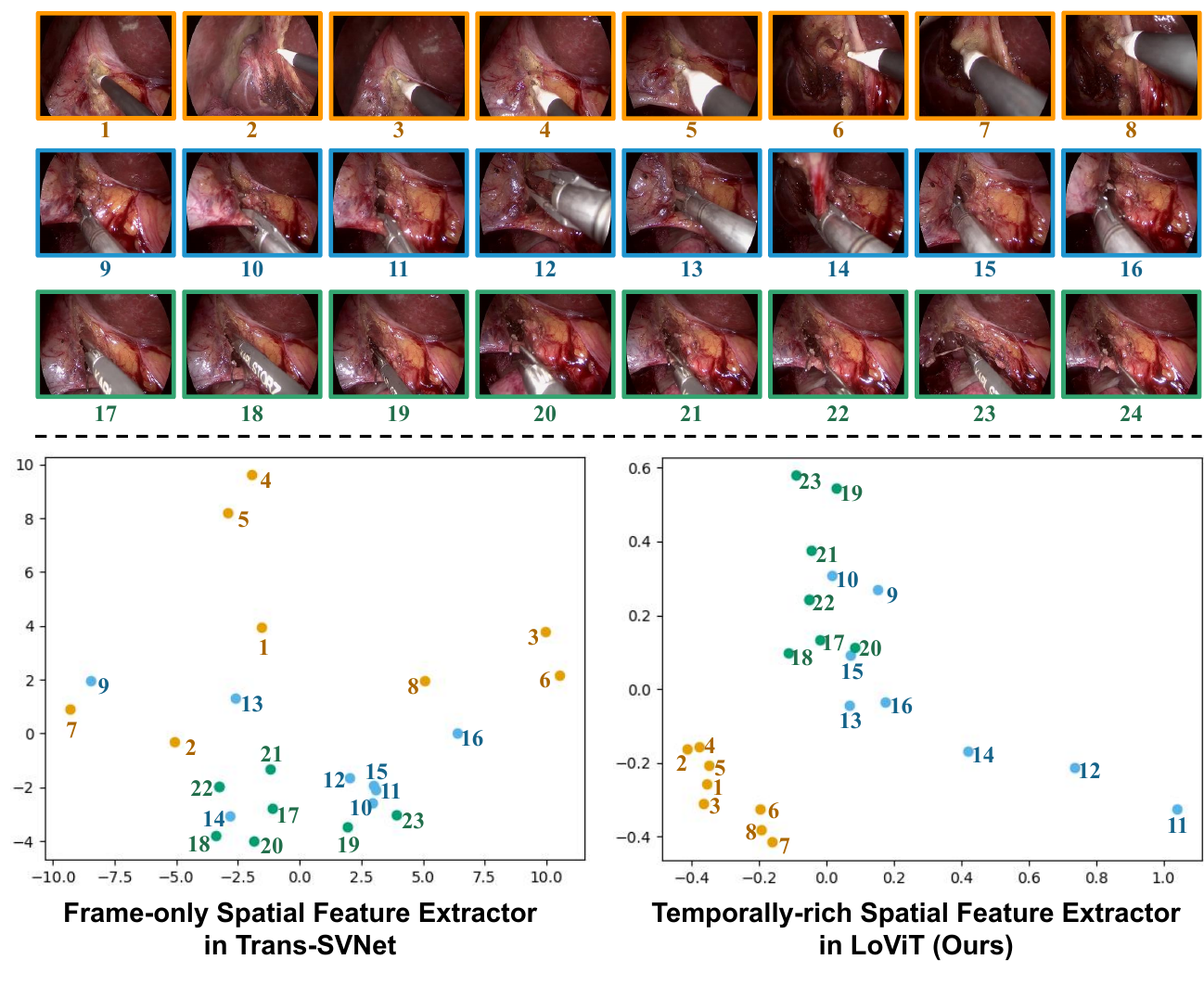}
\caption{Examples of spatial feature distribution of similar video frames.
\textit{Top:} three rows depict each frame that is similar, i.e. in regards to the tool environment. \textit{Bottom:} Visualization of the spatial feature distribution of example images using two different extractors.}
\label{fig:visual_spatial_examples}
\centering
\end{figure}
\subsection{Evaluating temporally-rich spatial feature extractor performance}
We investigated the impact of our temporally-rich spatial feature extractor. The results in Table~\ref{tab:sapcialextractor} demonstrate that using video clips as inputs for training temporally-rich spatial feature extractor contributes to higher performance, where the extractor inputs every single frame separately. This increase in performance is particularly notable on the AutoLaparo dataset, where we observe an improvement margin of 1.9 pp in accuracy. It is worth noting that even after replacing the temporally-rich spatial feature extractor with a normal spatial feature extractor, LoViT still outperforms current state-of-the-art methods 
To better understand the superiority of our temporally-rich spatial feature extractor compared to Trans-SVNet, we conducted a principal component analysis (PCA)~\cite{hotelling1933analysis} on their feature space and projected them into a two-dimensional reduced space representation for visualization. In Fig.~\ref{fig:visual_spatial_wholevideo}, we plotted the 2-dim feature representation of all frames in a video, labeled by the instruments appearing in each frame. We observed that our proposed temporally-rich spatial feature extractor was more effective in differentiating between different tools than the frame-only training method used in Trans-SVNet, as there was greater distinction between labeled frames in the reduced space. However, using tool labels to describe spatial features is insufficient for studying feature representations, since other objects appearing in a scene such as organs may also affect image features. To further investigate this, we manually selected three groups of illustrative frames with similar organ context and grouped them by the instrument appearing in the scene. We then visualized their spatial feature distributions in Fig.~\ref{fig:visual_spatial_examples}, providing further evidence of the superiority of our method over previous spatial feature extractors.

\subsection{Evaluating phase transition-aware supervision performance}

\begin{table}[t!]

\caption{Effects of adding phase transition-aware supervision on video- and phase-level metrics~(\%) when evaluated on Cholec80 and AutoLaparo datasets. Note that `$\checkmark$' means adding phase transition-aware supervision.}

\renewcommand\arraystretch{1.2}
\resizebox{\linewidth}{!}{
\begin{tabular}{cccccc}
\hline

\multirow{2}{*}{Dataset} &\multirow{2}{*}{Phase Transition-aware} &Video-level Metric  &\multicolumn{3}{c}{Phase-level Metric}\\\cline{3-6}
 & & $\mathrm{Accuracy}\uparrow$ &$\mathrm{Precision}\uparrow$ &$\mathrm{Recall}\uparrow$ &$\mathrm{Jaccard}\uparrow$\\\hline

%\multicolumn{2}{c|}{VI} & \multirow{2}{*}{Time (s)} \\ \cline{3-12}
%                                      && ODS         & OIS        & ODS        & OIS        & ODS        & OIS        & ODS         & OIS        & ODS         & OIS &\\\midrule[1pt]
\multirow{2}{*}{Cholec80} &           & $90.07 \pm 5.98$      & $82.23$        & $84.58$         & $71.67$\\
\cline{2-6}
&\checkmark            & $\bf{91.50 \pm 6.10}$      & $\bf{83.07}$       & $\bf{86.5}$        & $\bf{74.15}$\\ \hline

%\bottomrule[1pt] 
\multirow{2}{*}{AutoLaparo} &           & $77.86 \pm 7.88$      & $71.03$       & $64.78$        & $52.56$\\
\cline{2-6}
&\checkmark           & $\bf{81.43 \pm 7.35}$      & $\bf{85.07}$       & $\bf{65.85}$        & $\bf{55.90}$\\ \hline

%\bottomrule[1pt] 
\end{tabular}}

\label{tab:heatmap}
\end{table}

We evaluate the influence of our proposed heatmap for better learning phase transitions of surgical videos. 
Table~\ref{tab:heatmap} illustrates the improvement in our model's performance on both datasets when we used supervised phase transition map to capture the phase transition areas. These findings emphasize the importance of phase transition areas in surgical videos, as they contain critical information for identifying the start and end of each phase. By capturing these transitions, we gain a better understanding of phase relationships, which ultimately helps in reducing confusion between similar clips.

\subsection{Discussion}

Spatial feature extractor networks are typically trained for surgical phase recognition using image-only level supervision due to limited computing resources. However, our work indicates that a temporally-rich spatial feature extractor is essential for supporting an accurate recognition network.
In this study, we demonstrate that our temporally-rich spatial feature extractor on two datasets whilst having video clips as inputs is better than image-only level supervision for training a spatial feature extractor. 
This approach is particularly beneficial for surgical videos since there might be a large number of frames with similar characteristics across different phases due to some scenes changing only slightly and showing a limited amount of tools. 
We illustrate this in Fig.~\ref{fig:confusionfig} with scenes having similar spatial features and actions occasionally appearing at different phases~(classes), resulting in the extractor being confused and in our model to overfit because different labels supervise the same spatial features. 
Since key information in surgical videos is generally found at the beginning of every phase for distinguishing the phase of a given frame, we select a fix number of frame images from the beginning of the current phase to the current frame considered for classification spaced at regular intervals, whilst having the current frame's phase label as output to train our rich spatial feature extractor. By doing this, our model retains the key temporal information to recognize the phase of the current frame on the limitation of memory capacity. 

Similar to the above mentioned problem, similar actions occasionally appear at different surgical phases. Therefore, surgical phase recognition requires models that can process long sequences of video frames and extract their temporal relationships since interventions are typically long, generally lasting between one to two hours for cholecystectomy (Cholec80) and hysterectomy (AutoLaparo), whilst other interventions could even last longer, especially when complications arise.
Although the two most recent state-of-the-art methods for surgical phase recognition, namely TeCNO~\cite{DBLP:conf/miccai/CzempielPKSFKN20} and Trans-SVNet~\cite{DBLP:conf/miccai/GaoJLDH21}, are capable of processing long videos using TCNs~\cite{DBLP:conf/eccv/LeaVRH16}, their dilated temporal aggregation approach is unable to handle misclassifications of surgical phases due to the loss of fine-grained features. In this work, we adopted a Transformer-based model to aggregate temporal features, including vanilla self-attention mechanism for local short video clips and \textit{ProbSparse} self-attention mechanism for global long video, which outperforms TeCNO and Trans-SVNet. Moreover, even Trans-SVNet under-performed AVT in some cases as shown in Fig.~\ref{fig:visualize_comparewithsota}, a method that could only input short videos. Accordingly, from our results, it is suggested that dilated convolution operations that are part of TCN result in the model losing fine-grained features and continuous information.
In this way, vanilla and $ProbSparse$ self-attention operations demonstrate better performance over TCNs within the scope of our study.

Surgical videos are characterised by time dependency among phases, and pinpointing the phase transition areas is essential for discovering such dependency. To the best of our knowledge, we are the first to represent phase transitions with a phase transition map, which is utilised to supervise the model. From the experimental results, we observe that this phase transition map improves the performance of our proposed model. We notice that phase transition-aware supervision is an easy operation and does not impose an additional burden on the model.

Considering the experiment results on datasets Cholec80 and AutoLaparo, we observe that AutoLaparo is more challenging than Cholec80. Apart from the lower performances of all methods on AutoLaparo, we found that different previous methods performed similarly on it. However, LoViT outperformed TMRNet, TeCNO, and Trans-SVNet. Specifically, we proved that local information is more valuable than the global relationship in the experimental results. Most videos in AutoLaparo contain repeated phases, which causes more complex phase relationships than videos in Cholec80.
This limitation is exacerbated by the small sample size of AutoLaparo, which result in models having lower performance when compared to their performance observed in Cholec80.

\section{Conclusions}

We propose a new surgical phase recognition method named LoViT, which adopts video-clip level supervision to train a temporally-rich spatial feature extractor first and then uses the Multi-scale temporal feature aggregator to combine local fine-grained and global macroscopic information to recognise phases. Specifically, our Transformer-based LoViT allows long video feeds with less loss of information than other existing methods. Moreover, our LoViT utilises the heatmap to learn phases transition, which is significant to grab the relationship between phases. The proposed LoViT achieves state-of-the-art performance with great improvement over existing methods.

Even though LoViT shows superior performance, it remains difficult for LoViT to accurately recognise some phases that appear in an unusual operation process. As shown in the left video of Fig.~\ref{fig:visualize_comparewithsota}, `P5' appears after `P4' in most videos, but before `P4' in this video, which is difficult to be recognised by LoViT, and other recognition methods. We believe future research on online surgical phase recognition will focus more on discovering complex relationships between phases, an area we envisage continuing to investigate on.
Furthermore, despite the efficiency of LoViT's \textit{ProbSparse} self-attention mechanism, it still requires feeding all previous spatial features into the temporal model for recognition of each current frame. As the duration of the surgery increases, the inference speed of LoViT will deteriorate if it takes all video frames input. An ideal method, however, should avoid redundant calculations by utilizing the previous analysis results, thereby reducing time and memory costs, improving the stability of the system's inference speed, and enabling the processing of videos of any length, which aligns with our research direction.

{\small
\bibliographystyle{IEEEtran}
\bibliography{egbib}
}

\end{document}